\pdfoutput=1 

\documentclass[conference]{IEEEtran}

\ifCLASSINFOpdf
   \usepackage[pdftex]{graphicx}
\else

\fi

\usepackage{amsmath} 
\usepackage{adjustbox}

\usepackage[noadjust]{cite}

\begin{document}

\title{Machine learning approach in the development of building occupant personas}

\author{\IEEEauthorblockN{
Sheik Murad Hassan Anik\IEEEauthorrefmark{1},
Xinghua Gao\IEEEauthorrefmark{2} and
Na Meng\IEEEauthorrefmark{1}
}
\IEEEauthorblockA{\IEEEauthorrefmark{1}Department of Computer Science, Virginia Tech, Blacksburg, VA 24061, USA}
\IEEEauthorblockA{\IEEEauthorrefmark{2}Myers-Lawson School of Construction, Virginia Tech, Blacksburg, VA 24061, USA}}

\maketitle

\thispagestyle{plain}
\pagestyle{plain}

\begin{abstract}
The user persona is a communication tool for designers to generate a mental model that describes the archetype of users. Developing building occupant personas is proven to be an effective method for human-centered smart building design, which considers occupant comfort, behavior, and energy consumption. Optimization of building energy consumption also requires a deep understanding of occupants' preferences and behaviors. The current approaches to developing building occupant personas face a major obstruction of manual data processing and analysis. In this study, we propose and evaluate a machine learning-based semi-automated approach to generate building occupant personas. We investigate the 2015 Residential Energy Consumption Dataset with five machine learning techniques — Linear Discriminant Analysis, K Nearest Neighbors, Decision Tree (Random Forest), Support Vector Machine, and AdaBoost classifier — for the prediction of $16$ occupant characteristics, such as age, education, and, thermal comfort. The models achieve an average accuracy of $61\%$ and accuracy over $90\%$ for attributes including the number of occupants in the household, their age group, and preferred usage of heating or cooling equipment. The results of the study show the feasibility of using machine learning techniques for the development of building occupant persona to minimize human effort.
\end{abstract}

\IEEEpeerreviewmaketitle

\section{Introduction}


People spend most of the time indoors \cite{klepeis2001national}, and the impact of a building on its occupants is significant \cite{bell2003occupant}. Smart Buildings are buildings that integrate intelligence, enterprise, control, materials, and construction as an entire building system, with adaptability, not reactivity, at its core, to meet the drivers for building progression: energy and efficiency, longevity, and comfort and satisfaction \cite{buckman2014smart}. In recent years, human centeredness has become an increasingly important factor for smart building design and operation. Optimizing building performance requires a deep understanding of occupants’ behavior and preferences. Developing building occupant personas is proven to be effective in creating occupant profiles for human-centered smart buildings \cite{agee2021human}. The user persona is a communication tool for designers to generate a mental model that describes the archetype of users \cite{brangier2011persona}. The development of a building occupant persona can assist the building simulation community in more accurately estimating energy demand through realistic and representative occupant profiles, enabling pragmatic occupant-centric building design and operation. The increasing availability of data from a wide range of sources will allow smart buildings to become adaptable, and prepare themselves for context and change over all timescales \cite{buckman2014smart}. 

One of smart buildings' key objectives is to maximize occupant comfort while minimizing energy consumption \cite{mo2002intelligent}. Buildings and urban spaces increasingly incorporate artificial intelligence and new forms of interactivity, raising a wide span of research questions about the future of human experiences with, and within, built environments. This emerging area is defined as Human-Building Interaction (HBI) \cite{alavi2019introduction}. HBI affects human well-being and the surrounding environment. Although people place energy and technology into buildings for their own comfort and recreation, technology often leads to the decision-making process excluding the human factor from the design itself. Smart buildings provide an opportunity to design for the physical, physiological, and psychological needs of occupant-users. Human-centered design (HCD) places the human at the center of the building design, which can accelerate smart housing design for people. To maximize human well-being and the optimize performance of smart buildings, an iterative, human-centered approach to building design must be employed.
\\

Agee \textit{et al}. \cite{agee2021human} introduced a human-centered approach to smart housing, leading to the development of data-driven smart housing personas that communicate smart housing user needs. In their work, the authors utilized descriptive statistics and behavioral analysis to describe the physical, physiological, and psychological needs of the occupants. Personas help designers anchor their work in a fictional user’s needs in the design of products and systems \cite{takai2010use}. This approach reduces the risk of designers designing for themselves, technology, and/or first cost parameters. 
\\


Developing building occupant personas benefits both the occupants and building designers in many ways. However, the task faces multiple challenges beginning with the limited availability of building occupant data. The lack of data related to occupants and their living environment is one of the major pullbacks in smart housing persona development research. Even with the limited data that is available, the process of developing building occupant persona is manual and time-consuming \cite{agee2021human}. Currently, the entire occupant persona development process requires researchers manually analyzing data, conducting interviews, clustering the occupant profiles based on multiple criteria, and finally constructing a persona. The entire process takes much time because it involves manual labor in all the steps. Automating a single step of this procedure can accelerate the entire persona development task. Moreover, although there is a scarcity of residential building data, there has been some work to address this issue recently. For example, Anik \textit{et al}. \cite{anik2022cost} presented a cost effective and portable framework for indoor data collection, and Song, \textit{et al}. \cite{song2019human} discussed data collection and analysis methods for analyzing human comfort in the indoor environment. With more building data available, and the development of machine learning technologies, new opportunities are emerging. Machine learning is an automated process that extracts patterns from data \cite{kelleher2020fundamentals}. In the field of predictive data analytics, machine learning is a method used to devise complex prediction algorithms and models \cite{mitchell1997artificial}. Machine learning models that can deliver fast and accurate results given enough data are provided \cite{pantic2005introduction}. 
\\

In this work, we apply machine learning to the 2015 Residential Energy Consumption Survey Data to automate some steps of building occupant persona development. The procedure includes (i) data processing, (ii) feature engineering, (iii) selection of machine learning models, (iv) selection of target variables, (v) selection of descriptive variables, (vi) training the models, and (vii) evaluating the models on unseen data. This work aims to answer the following research questions: 
\\
\begin{itemize}
    \item RQ1: Is it possible to automate the process or some steps of building occupant persona development?
    \item RQ2: How accurate the results can be achieved in the automated process of building occupant persona development? 
\end{itemize}

The rest of the paper is structured as follows: Section 2 reviews relevant studies conducted in the domain of building occupant persona development and machine learning methods for occupant behavior modeling. Section 3 presents the methodology, description of the data, and machine learning models used. Section 4 demonstrates the machine learning results. Section 5 discusses the findings, the answers to the research questions, the limitations, and the future research directions. Section 7 concludes the research.


\section{Related Works}
Related previous studies can be grouped into the following two categories: building occupant persona development and using machine learning to model occupant behavior.

\subsection{Building occupant persona development}


 HBI usually reflects some characteristics of the building occupants, and the building occupant persona is an effective and concise way to represent these characteristics for human-centered building design and operation. Agee \textit{et al}. \cite{agee2021human} conducted a study on a human-centered approach to smart housing. The study employed a multi-phase, mixed-methods research design and collected data from 309 residential housing units in Virginia, U.S. The authors collected longitudinal energy use data via the deployed electricity sensors and the occupants' attitudes, preferences, and self-perceived energy behaviors via surveys and semi-structured interviews. Affinity diagramming was then used to identify the topics and themes of occupants' HBI activities. The output of the affinity diagramming analysis and energy analysis led to the development of data-driven Personas that communicate smart housing user needs. Malik \textit{et al}. \cite{malik2022developing} researched the occupant behavior of 1223 low-income households in India through a transverse field study. Three occupant archetypes, indifferent consumer, considerate saver, and conscious conventional, were established based on the occupants’ behavioral and psychographic characteristics gathered from field surveys. Occupants’ age and education were determined as significant factors influencing occupant archetype clustering and their actions within the indoor built environment. The segmentation results further inform that there exists considerable diversity in occupant behavior and attitudes within similar socio-economic groups, which consequently leads to differential energy demand and occupant comfort preferences. The findings can assist the building simulation community in accurately estimating energy demand through realistic occupant profiles and providing pragmatic occupant-centric building designs for the future low-income housing stock.
 \\
 
Studying occupant behavior related to their comfort needs is critical for better understanding home energy consumption. Ortiz \textit{et al}. \cite{ortiz2019developing} studied the motivations behind comfort behaviors and the energy consumption discrepancies among occupants with different behavioral patterns. The authors grouped the occupants into five categories based on their psychological and behavioral models --- locus of control, emotions towards their home environment, and the importance they give to comfort affordances. The findings show that each of the archetypes has a distinct valence of opinions when asked about topics regarding energy use, energy awareness, general comfort, and affordances, but what they express verbally is not always congruent to the general results of their self-reported answers. Buttitta \textit{et al}. \cite{buttitta2019development} proposed a method to develop occupancy-integrated archetypes that allow the annual final heating energy required (by building stock) characterized by different occupancy profiles, which play a significant role in influencing heat demand in residential buildings. The authors showed that, for UK residential archetypes, the discrepancy between the heat demand calculated using the proposed occupancy-integrated archetypes and the BRE Domestic Energy Model (BREDEM) calculation procedures can be up to 30\%, pointing out that the use of BREDEM occupancy profiles is not necessarily appropriate, particularly when disaggregated and differentiated energy profiles are involved.
\\

\subsection{Using machine learning to model occupant behavior}

Occupant behavior remains one of the main sources of uncertainty in building energy modeling \cite{demodeling}. Amasyali \textit{et al}. \cite{amasyali2021real} proposed a data-driven method of modeling occupant behavior for simultaneously reducing energy consumption and improving comfort. The proposed method consists of two components: 1) a set of machine learning-based occupant-behavior-sensitive models for predicting energy consumption and thermal and visual comfort, and 2) a genetic algorithm-based optimization model for optimizing occupant behavior. To test and evaluate the proposed method, an office building was instrumented and data about energy consumption, outdoor weather conditions, occupant behavior, and occupant comfort were collected for about three months. Based on the collected data, the authors developed a set of machine learning models for predicting energy consumption of cooling and lighting and occupant comfort of thermal and visual. Carlucci \textit{et al}. \cite{carlucci2020modeling} presented a similar study with a machine learning approach for predicting building energy consumption in an occupant-behavior-sensitive manner. In their approach, a model learns from a large set of energy-use cases that were modeled and simulated in EnergyPlus \cite{crawley2001energyplus}. The machine-learning prediction model was trained using a large dataset including 3-month hourly data for 5760 use cases representing different combinations of building characteristics, outdoor weather conditions, and occupant behaviors. The results of the study demonstrated the high impact of occupant behavior on building energy consumption and identified opportunities for behavioral energy-saving measures.
\\

To create comfortable indoor environments for building occupants, Deng \textit{et al}. \cite{deng2018artificial} developed artificial neural network (ANN) models for predicting indoor thermal comfort by using thermal sensations and occupants’ behavior. The models were trained by data about air temperature, relative humidity, clothing insulation, metabolic rate, thermal sensations, and occupants’ behavior collected in ten offices and ten houses/apartments. The models were able to predict similar acceptable air temperature ranges in offices. The occupants’ behavior in multi-occupant offices was more complex, which led to a different prediction of thermal comfort. This investigation demonstrates alternative approaches to the prediction of thermal comfort.
\\

Peng \textit{et al}. \cite{peng2018using} conducted an analysis of occupants’ behavior in an office building with the goal of increasing the efficiency of HVAC systems. They proposed a demand-driven control strategy that automatically responds to occupants’ energy related behavior for reducing energy consumption and maintains room temperature for occupants with similar performances. In this control strategy, two types of machine learning methods, unsupervised and supervised learning, were applied to learn occupants’ behavior in two learning processes. The occupancy related information learned by the algorithms was used by a set of specified rules to infer real-time room set-points for controlling the office's space cooling system. This learning-based approach intends to reduce the need for human intervention in the cooling system’s control. The proposed strategy was applied to control the cooling system of the office building under real-world conditions. 
\\

Automating the development of a human-centered building occupant persona requires vast data on both energy consumption and occupant behavior. The gap in either one not only hinders the comprehensiveness of the building occupant persona development process but also leads to inaccuracy in results. Although occupant behavior is difficult to model due to the stochastic nature and variability of humans, it is necessary to explore the generic pattern of their behaviors and integrate the information with the building energy model \cite{jia2018framework}. Modeling occupant behavior remains one of the key steps in the process of building occupant persona development. A valid occupant behavior model needs to have the potential to simulate realistic building users’ reactions to different built environments, and the lack of real occupant behavioral data accounts for a gap in research. The human-centered building occupant persona development process proposed in \cite{agee2021human} requires manual human work of labeling, filtering, and clustering data from large datasets. These steps are both labor and time extensive, which impede the entire process. To address these research gaps, this research utilizes real-world occupant data with machine learning tools to accelerate the process of building occupant persona development. 



\section{Methodology}
\label{sec:methodology}

\begin{figure}[ht]
\centering
\includegraphics[width=\columnwidth]{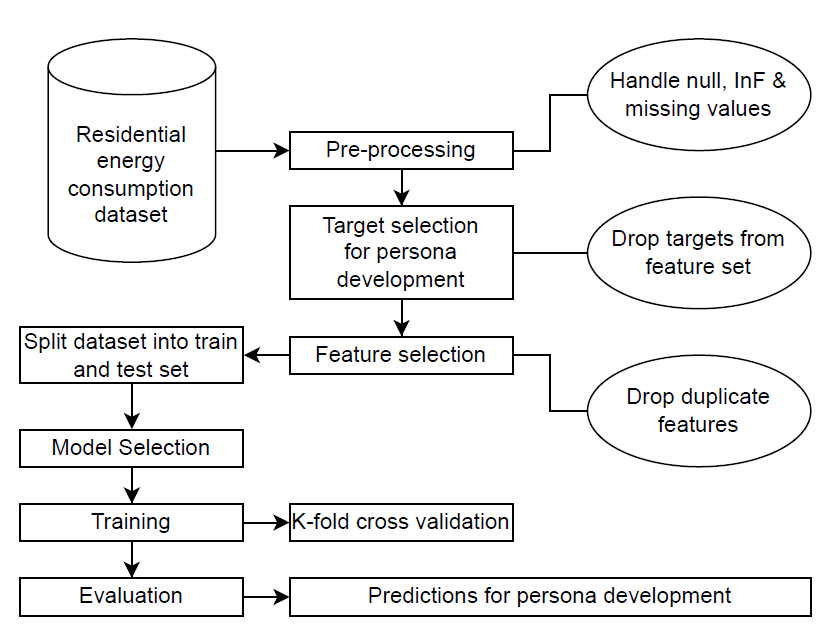}
\caption{Methodology to occupant characteristics prediction and building occupant persona development}
\label{fig:methodology}
\end{figure}

Figure \ref{fig:methodology} illustrates the framework of this research, which is inspired by a study conducted by Zhongguo \textit{et al}. \cite{zhongguo2017choosing}. The process begins with processing the 2015 Residential Energy Consumption Dataset \cite{eia_2018}. The metadata is extracted, and the data is cleaned in the pre-processing step. Then, $16$ target variables are chosen to represent occupant characteristics, such as age, education, income, thermal preference, etc. The rest of the attributes remain as descriptive variables and are filtered through the feature selection step, in which the irrelevant and redundant variables are dropped. The dataset is randomly split into training and testing sets following the $80-20$ rule \cite{seiler2010best}. Then, five machine learning models for classification prediction tasks are selected and tested. These models are trained through the 10-fold cross validation. After that, the generated models are evaluated on the testing set. 

\subsection{Dataset description}
The Residential Energy Consumption Survey (RECS) is a periodic study conducted by the U.S. Energy Information Administration (EIA) that provides detailed information about energy usage in U.S. homes \cite{eia_2018}. RECS is a multi-year effort consisting of the household survey, data collection from household energy suppliers, and end-use consumption and expenditures estimation. The household survey collects data on energy-related characteristics and usage patterns of a national representative sample of housing units. The energy supplier survey collects data on how much electricity, natural gas, propane/LPG, fuel oil, and kerosene were consumed in the sampled housing units during the reference year. RECS samples homes occupied as primary residences, which excludes secondary homes, vacant units, military barracks, and common areas in apartment buildings. As a result, RECS estimates do not represent sector-level totals but they are best suited for comparison across different characteristics of homes within the residential sector.
\\

The total number of responding households is $5686$ in the 2015 RECS. Each record includes a total of $759$ attributes. These attributes have been categorized in $12$ sections by EIA: 

\begin{itemize}
    \item A: Structural Characteristics -- house type, construction time, renovation time, floor plan, building materials, etc. 
    \item B: Kitchen Appliances -- kitchen stove, microwave, stove fuels, oven, hood, ventilation, grill, refrigerator, freezer, dishwasher, etc.
    \item C: Home Appliances and Electronics -- clothes washer, dryer, washing cycle, television, gaming console, phone, computer, etc. 
    \item D: Space Heating -- heating appliances, heater usage during winter, heating fuel, thermostat controls, hot water system, fireplace, etc. 
    \item E: Air Conditioning -- air conditioners, cooling systems, air conditioning heat pump, air filter, programmable thermostat, thermostat control during summer days, etc. 
    \item F: Water Heating -- water heaters types, fuels, usage, age, tank, etc. 
    \item G: Miscellaneous -- light bulbs, swimming pool, energy audit, energy assistance, etc.
    \item H: Fuels Used -- back-up generator, onsite power system details, fuel usage in home, energy bills, natural gas usage in home, etc. 
    \item I: Housing Unit Measurement -- size, shape and area of the different floors of the housing units.
    \item J: Fuel Bills -- fuel suppliers, bill types, different fuel bills like electricity, gas, wood, smart meter, etc. 
    \item K: Housing Unit Characteristics -- occupants' age, gender, education, employment status, number of people living in the household, income, etc. 
    \item L: Energy Insecurity and Assistance -- challenges paying bills, struggles due to unsafe or unhealthy temperature, utility discontinuation, energy assistance, requirement of medication, etc. 
\end{itemize}

\begin{figure}[ht]
\centering
\includegraphics[width=\columnwidth]{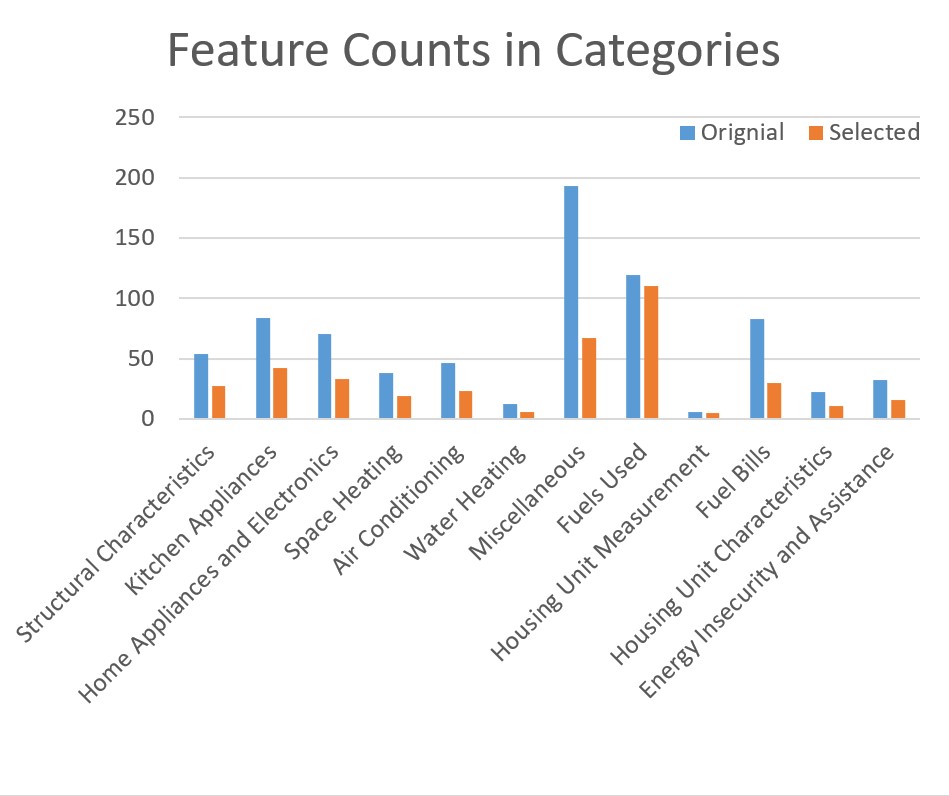}
\caption{Feature distribution in categories.}
\label{fig:combined_feature_dist}
\end{figure}

Figure \ref{fig:combined_feature_dist} illustrates the number of features in each of the $12$ categories in pairs. For every pair, the left bar (colored in blue) refers to the original number of features in that category and the right bar (colored in orange) refers to the number of features selected after the feature selection process. $16$ of these attributes are chosen to be the target variables because they relate to individual occupant characteristics. The rest of the attributes remain as input variables. The next section discusses the feature selection process in detail. The $5686$ records are divided randomly into training and testing sets by the $80-20$ rule \cite{seiler2010best}. In the training process, the k-fold cross validation is utilized with $k = 10$.  

\subsection{Pre-processing}
Data pre-processing is the process of transforming raw data into a machine-understandable format. In this work, the data of RECS 2015 is almost ready to use. However, some records contain infinite, missing, blank or null values. For machine learning models to understand all records, these values needed to be transformed. The infinite values are replaced with a large enough number, i.e. the largest integer. The blank or null values are replaced with $-1$ to establish the value is missing. One target variable, HHAGE which refers to the age of the occupant, is continuous from $18$ to $110$. It is difficult to fit this variable in the classification models. Studies \cite{yarlagadda2015novel, lin2020establishment} used age groups to tackle this issue in the past. This study follows the age range used in \cite{lin2020establishment}. The records are categorized in the following age groups (in years): Children (0 to 12), Young Adult (13 to 30), Middle Adult (31 to 50), Senior Adult(51 to 70), and Senior (71 to 110). 

\subsection{Feature selection}
Handling all $759$ features is difficult and makes the machine learning process slow, which leads to a feature selection process. After manual examination of each of the $759$ features, the authors drop $370$ features and use $389$ features in the model training. The original dataset included imputation flags for most of the attributes which refer to whether a record was imputed or not. A record can be imputed if it is not directly obtained from the occupant, but it is observed from other sources such as the structure, surrounding, or measuring device. Either way, in this work, the imputation flag is not a necessary component. Hence, the imputation flags are dropped. The energy consumption attributes recorded in the dataset contained both energy units and corresponding costs in US dollars. The dollar amount is not in the scope of this study, and thus is dropped from the dataset. The dropped $370$ features include imputation flags, utility bills in US dollar amount (actual unit is kept), and replicated weights (used for variance estimation). These are safe to drop because they do not represent any data that might affect the output of the machine learning models for developing building occupant personas. The attribute related to the number of phones is also removed because it directly correlates to the number of occupants or adults living in the house. The orange bars in Figure \ref{fig:combined_feature_dist} (right bar of each category pair) represent the number of features selected for the training models.

\subsection{Target variables}
\label{sec:target_variables}
We examine remaining $389$ attributes to select the target variables, and find $16$ columns related to occupant characteristics. These attributes can provide key information in developing building occupant persona: 

\begin{itemize}
    \item EQUIPMUSE: Main heating equipment household behavior, including values such as setting one temperature and leaving it there most of the time; manually adjusting the temperature at night or when no one is at home; programming the thermostat to automatically adjust the temperature during the day and night at certain times; turn equipment on or off as needed, etc. 
    \item TEMPHOME: Winter temperature when someone is home during the day with the value range of $50$ to $90$ degrees Fahrenheit.  
    \item TEMPGONE: Winter temperature when no one is home during the day with the value range of $50$ to $90$ degrees Fahrenheit.  
    \item TEMPNITE: Winter temperature at night with the value range of $50$ to $90$ degrees Fahrenheit.  
    \item USEWWAC: Most-used individual air conditioning unit household behavior including values, such as setting one temperature and leave it there most of the time; manually adjust the temperature at night or when no one is at home; programming the thermostat to automatically adjust the temperature during the day and night at certain times; turn equipment on or off as needed, etc. 
    \item TEMPHOMEAC: Summer temperature when someone is home during the day with the value range of $50$ to $90$ degrees Fahrenheit.  
    \item TEMPGONEAC: Summer temperature when no one is home during the day with the value range of $50$ to $90$ degrees Fahrenheit.  
    \item TEMPNITEAC: Summer temperature at night with the value range of $50$ to $90$ degrees Fahrenheit.  
    \item HHAGE: Respondent age, values ranging from $18$ to $110$.
    \item EMPLOYHH: Respondent employment status. Values covering employed full-time, part-time and unemployed or retired. 
    \item EDUCATION: Highest education completed by respondent. Replies cover less than high school diploma or GED, high school diploma or GED, some college or associate’s degree, bachelor’s degree (for example: BA, BS), master’s, professional, or doctorate degree (for example: MA, MS, MBA, MD, JD, PhD).
    \item NHSLDMEM: Number of household members, values ranging from $1$ to $20$.
    \item NUMADULT: Number of household members age 18 or older, values ranging from $1$ to $20$.
    \item NUMCHILD: Number of household members age 17 or younger, values ranging from $1$ to $20$.
    \item ATHOME: Number of weekdays someone is at home, values ranging from $1$ to $5$.
    \item MONEYPY: Annual gross household income for the last year. Replies cover a range less than $20000$ to more than $140000$ in $8$ sections. 
\end{itemize}

\begin{figure*}[ht]
\centering
\includegraphics[width=\textwidth]{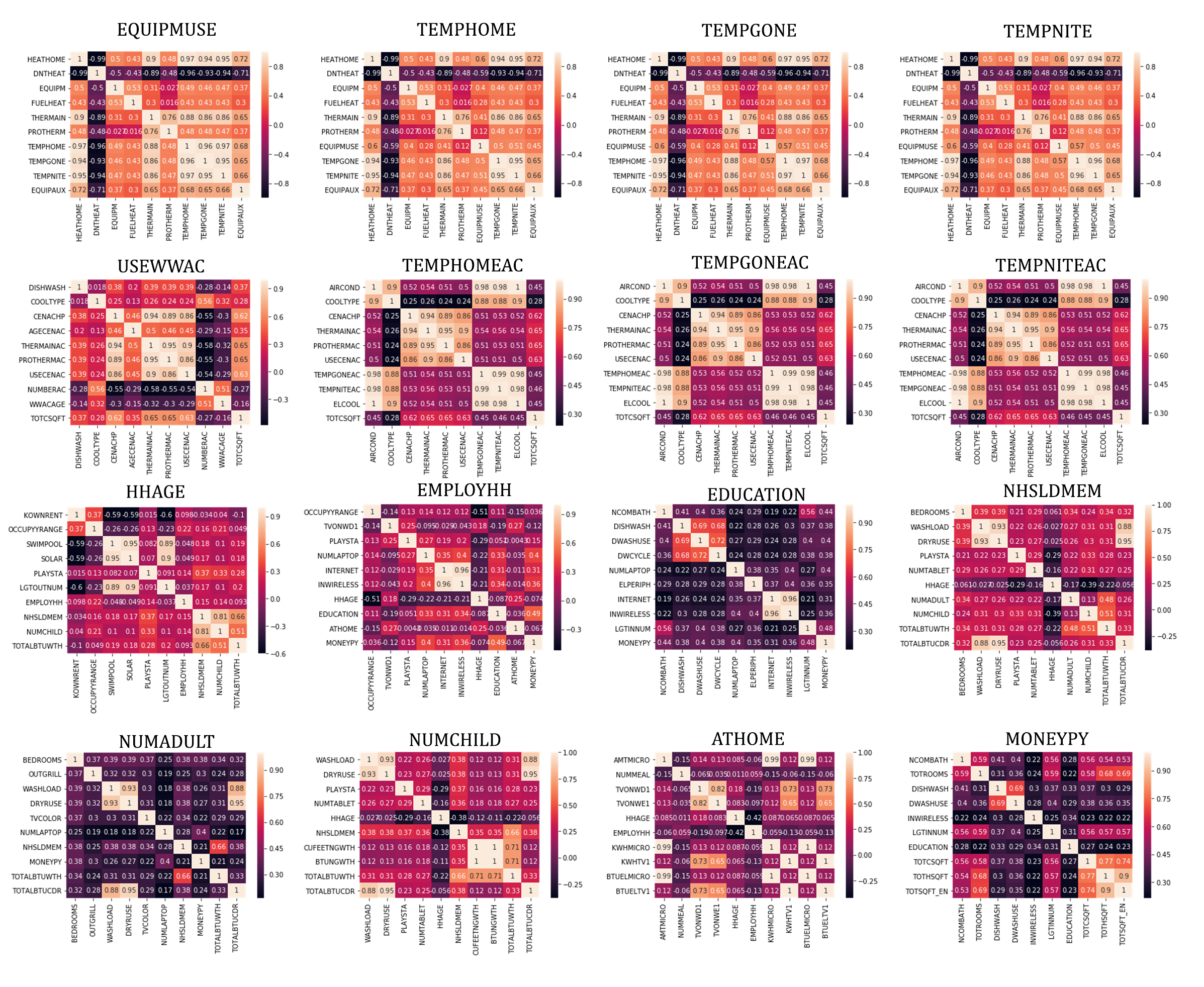}
\caption{Small Correlation Matrix for Each Target Variables.}
\label{fig:correlation}
\end{figure*}

The authors examine the correlations among the $389$ attributes and create correlation matrices. Figure \ref{fig:correlation} illustrates $16$ smaller correlation matrices for the $16$ target variables. Each of them is constructed with 10 most correlating attributes to respective target variables. For example, the most correlating attributes to the target EDUCATION are NCOMBATH, DISHWASH, DWASHUSE, DWCYCLE, NUMLAPTOP, ELPERIPH, INTERNET, INWIRELESS, LGTINNUM, and MONEYPY. The correlation figures are color coded, from darkest to lightest representing negative correlation to positive correlation. It is intuitive that the number of laptops, internet devices, and household income are related to the level of education of the occupant. On the contrary, it is interesting to find that the dish-washing cycle usage have a strong correlation with the level of education. The first row of matrices consists mostly of positive correlations whereas quite a few strong negative correlation is observed on the matrices of EDUCATION, NUMADULT, and MONEYPY. Matrices for USEWWAC, HHAGE, EMPLOYHH, and NHSLDMEM maintain a balance between the positive and negative correlation.

\subsection{Machine learning models}
Classification is a supervised learning approach in which a target variable is categorical or discrete. The task of choosing a specific classification model is a critical step \cite{kotsiantis2006machine}, and each model has its own strengths and weaknesses in a given scenario. There is no cut-and-dried flowchart that can be used to determine which model should be used or will outperform the rest. A simple example can be, back propagation neural networks achieve higher accuracy than the decision tree method on Iris and Appendicitis data but a lower accuracy on Breast cancer and Thyroid data \cite{weiss1989empirical, zhongguo2017choosing}. Pedregosa \textit{et al.} \cite{scikit-learn} suggests a guide on choosing the proper machine learning algorithm for a given task. In this work, five machine learning models have been used parallelly on their default configuration.  The following are the classification models used in this work: 

\begin{itemize}
    \item Linear Discriminant Analysis (LDA): It is a linear model for classification and is most commonly used for feature extraction in pattern classification problems. It is used in finding the projection hyper-plane that minimizes the inter-class variance and maximizes the distance between the projected means of the classes \cite{xanthopoulos2013linear}. Logistic Regression is one of the most popular linear classification models that perform well for binary classification but falls short in the case of multiple classification problems with well-separated classes. While LDA handles these quite efficiently.
    \item K Nearest Neighbors Classifier (KNN): The nearest neighbor (NN) classifiers, especially the k-NN algorithm, are among the simplest and yet most efficient classification rules and are widely used in practice \cite{laaksonen1996classification}. It’s easy to implement and understand. KNN works by finding the distances between a query and all the examples in the data, selecting the specified number of examples (K) closest to the query, then voting for the most frequent label (in the case of classification).
    \\
    \item Decision Tree Classifier (CART): A Decision Tree is a simple representation for classifying examples. It is a Supervised Machine Learning where the data is continuously split according to a certain parameter \cite{swain1977decision}. The key idea is to use a decision tree to partition the data space into cluster (or dense) regions and empty (or sparse) regions. In Decision Tree Classification a new example is classified by submitting it to a series of tests that determine the class label of the example. These tests are organized in a hierarchical structure called a decision tree. Decision Trees follow Divide-and-Conquer Algorithm.
    \\
    \item Support Vector Machine (SVM): The support vector machine is a machine learning method based on statistical learning theory. Because of its higher accuracy, SVM has become the research focus of the machine learning community \cite{zhang2012support}. In the SVM algorithm, each data is plotted as a point in N-dimensional space (where N is the number of features) with the value of each feature being the value of a particular coordinate. Then, classification is performed by finding the hyper-plane that differentiates the two classes very well. The objective of the support vector machine algorithm is to find a hyperplane in the N-dimensional space that distinctly classifies the data points.
    \\
    \item AdaBoost Classifier (ADB): AdaBoost is an ensemble learning method (also known as “meta-learning”) that was initially created to increase the efficiency of binary classifiers. AdaBoost uses an iterative approach to learn from the mistakes of weak classifiers, and turn them into strong ones \cite{Freund1996ExperimentsWA, schapire2013explaining}. A single classifier may not be able to accurately predict the class of an object, but when multiple weak classifiers are grouped with each one progressively learning from the others' wrongly classified objects, a strong model can be built.
    \\
    \item Random Forest Classifier (RFC): Random forest is a supervised learning algorithm. The "forest" it builds, is an ensemble of decision trees, usually trained with the “bagging” method. The general idea of the bagging method is that a combination of learning models increases the overall result \cite{parmar2018review}. Random forest is a flexible, easy-to-use machine learning algorithm that produces, even without hyper-parameter tuning, a great result most of the time. It is also one of the most used algorithms, because of its simplicity and diversity. 
\end{itemize}


\section{Results}
\label{sec:results}

\begin{figure*}[ht]
\centering
\includegraphics[width=\textwidth]{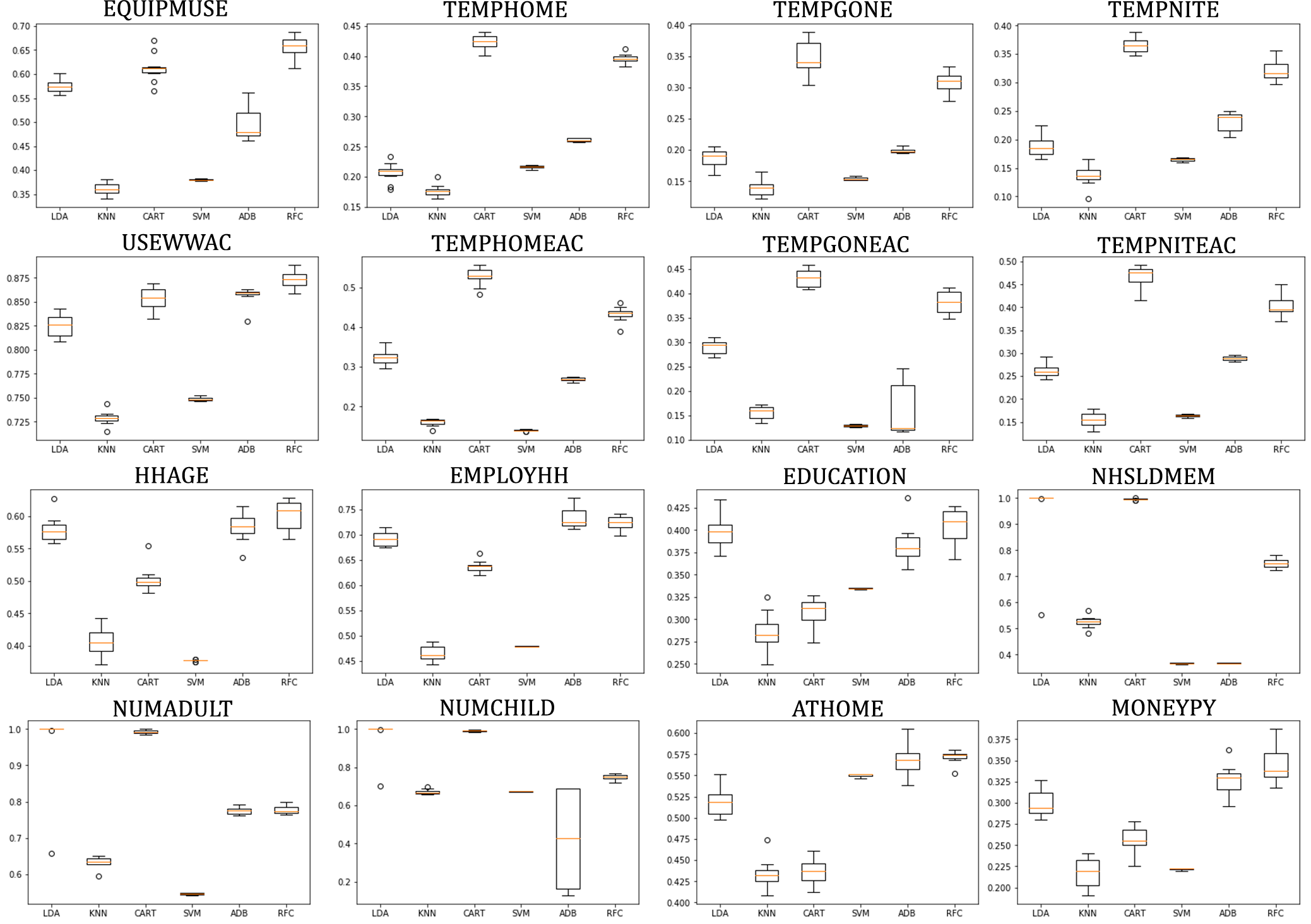}
\caption{Training Accuracy over multiple iteration}
\label{fig:accuracy_spread}
\end{figure*}

Figure \ref{fig:accuracy_spread} illustrates the accuracy obtained for all $16$ target variables during the 10-fold cross-validation training process. $80\%$ of the dataset is used for the training process. In Figure \ref{fig:accuracy_spread}, the training models are lined from left to right in the x-axis of each figure in the order: Linear Discriminant Analysis (LDA), K Nearest Neighbor (KNN), Decision Tree Classifier (CART) Support Vector Machine (SVM), AdaBoost Classifier (ADB), and Random Forest Classifier (RFC). All models have been used with their default parameters without any hyper tuning. 

\begin{table}[]
\caption{Training accuracy of target variables}
\begin{tabular}{lllllll}
 & LDA & KNN & CART & SVM & ADB & RFC \\ \hline
EQUIPMUSE & 0.57 & 0.36 & 0.61 & 0.38 & 0.49 & 0.66 \\ \hline
TEMPHOME & 0.21 & 0.18 & 0.42 & 0.21 & 0.26 & 0.4 \\ \hline
TEMPGONE & 0.19 & 0.14 & 0.35 & 0.15 & 0.2 & 0.31 \\ \hline
TEMPNITE & 0.19 & 0.14 & 0.37 & 0.16 & 0.23 & 0.32 \\ \hline
USEWWAC & 0.83 & 0.73 & 0.85 & 0.75 & 0.86 & \textbf{0.87} \\ \hline
TEMPHOMEAC & 0.32 & 0.16 & 0.53 & 0.14 & 0.27 & 0.43 \\ \hline
TEMPGONEAC & 0.29 & 0.16 & 0.43 & 0.13 & 0.15 & 0.38 \\ \hline
TEMPNITEAC & 0.26 & 0.15 & 0.47 & 0.16 & 0.29 & 0.4 \\ \hline
HHAGE & 0.58 & 0.41 & 0.51 & 0.38 & 0.58 & 0.6 \\ \hline
EMPLOYHH & 0.69 & 0.47 & 0.64 & 0.48 & 0.73 & 0.72 \\ \hline
EDUCATION & 0.4 & 0.28 & 0.31 & 0.33 & 0.38 & 0.41 \\ \hline
NHSLDMEM & 0.95 & 0.53 & \textbf{0.995} & 0.37 & 0.37 & 0.75 \\ \hline
NUMADULT & 0.97 & 0.63 & \textbf{0.99} & 0.55 & 0.78 & 0.77 \\ \hline
NUMCHILD & 0.97 & 0.67 & \textbf{0.99} & 0.67 & 0.42 & 0.75 \\ \hline
ATHOME & 0.51 & 0.43 & 0.44 & 0.55 & 0.57 & 0.57 \\ \hline
MONEYPY & 0.3 & 0.21 & 0.26 & 0.22 & 0.33 & 0.34
\end{tabular}
\label{tab:train_acc}
\end{table}

\begin{table}[]
\caption{Test accuracy of target variables}
\begin{tabular}{lllllll}
 & LDA & KNN & CART & SVM & ADB & RFC \\ \hline
EQUIPMUSE & 0.59 & 0.37 & 0.61 & 0.38 & 0.48 & 0.67 \\ \hline
TEMPHOME & 0.21 & 0.17 & 0.43 & 0.21 & 0.26 & 0.37 \\ \hline
TEMPGONE & 0.19 & 0.13 & 0.38 & 0.14 & 0.19 & 0.34 \\ \hline
TEMPNITE & 0.2 & 0.13 & 0.36 & 0.16 & 0.24 & 0.33 \\ \hline
USEWWAC & 0.85 & 0.76 & 0.86 & 0.78 & 0.87 & \textbf{0.89} \\ \hline
TEMPHOMEAC & 0.31 & 0.17 & 0.53 & 0.14 & 0.27 & 0.44 \\ \hline
TEMPGONEAC & 0.28 & 0.14 & 0.42 & 0.13 & 0.11 & 0.37 \\ \hline
TEMPNITEAC & 0.28 & 0.16 & 0.46 & 0.16 & 0.29 & 0.41 \\ \hline
HHAGE & 0.61 & 0.42 & 0.49 & 0.38 & 0.6 & 0.64 \\ \hline
EMPLOYHH & 0.68 & 0.49 & 0.64 & 0.47 & 0.73 & 0.72 \\ \hline
EDUCATION & 0.37 & 0.26 & 0.31 & 0.33 & 0.36 & 0.39 \\ \hline
NHSLDMEM & \textbf{1} & 0.53 & 0.99 & 0.38 & 0.38 & 0.78 \\ \hline
NUMADULT & \textbf{1} & 0.64 & 0.99 & 0.54 & 0.76 & 0.77 \\ \hline
NUMCHILD & \textbf{1} & 0.68 & 0.99 & 0.69 & 0.69 & 0.75 \\ \hline
ATHOME & 0.51 & 0.46 & 0.45 & 0.57 & 0.57 & 0.57 \\ \hline
MONEYPY & 0.31 & 0.21 & 0.28 & 0.22 & 0.34 & 0.33
\end{tabular}
\label{tab:test_acc}
\end{table}

Classification accuracy of the developed models on unseen data is one of the principal metrics for evaluating classification models. The following formula can be used to determine classification accuracy: 

\begin{equation}
    Accuracy = \frac{Number Of Correct Predictions}{Total Number Of Predictions}
\label{eq:accuracy_eq}
\end{equation}

Table \ref{tab:train_acc} reveals the average accuracy obtained during the training process. The classification models obtain above $75\%$ training accuracy for $4$ out of $16$ target variables and above $50\%$ for $9$ out of  $16$ target variables. Some of the models obtain below $25\%$ training accuracy for $7$ out of $16$ target variables. The models perform exceptionally well and achieve $99\%$ to $100\%$ accurate for $3$ target variables, EDUCATION, NHSLDMEM, and NUMADULT. Based on the results shown in Table \ref{tab:train_acc} and Figure \ref{fig:accuracy_spread}, the Decision Tree Classifier (CART) outperforms the other classifiers used in this task. 
\\\\

Table \ref{tab:test_acc} illustrates the accuracy obtained during the test process. During the testing set evaluation, the $20\%$ unknown portion of the dataset is used to capture unbiased results. The train and test accuracy graphs show that the testing results closely imitate the training accuracy curves. The trained machine learning models retain above $75\%$ test accuracy for $4$ out of $16$ target variables and above $50\%$ for $9$ out of $16$ target variables. Similar to the training accuracy, the trained models perform $99\%$ to $100\%$ accurately for EDUCATION, NHSLDMEM, and NUMADULT on the testing set. Overall, in terms of accuracy, the Decision Tree Classifier (CART) outperforms the other classifiers used in this task. The ensemble models, AdaBoost and Random Forest Classifiers maintained a near-constant performance thought out different targets. 
\\

For some of the target variables, the accuracy is significantly lower than the rest. For example, the prediction accuracy temperature during different times obtained lower accuracy (less than $50\%$). The reason can be two-fold. The first one is the wrong model selection for the task and the second one is the absence of hyper-tuning for specific target variables.

\begin{table}[]
\caption{Mean absolute error (MAE) in test set}
\begin{tabular}{lllllll}
 & LDA & KNN & CART & SVM & ADB & RFC \\ \hline
EQUIPMUSE & 0.63 & 1.15 & 0.67 & 1.23 & 0.79 & 0.57 \\ \hline
TEMPHOME & 2.98 & 7.89 & 2.29 & 6.32 & 1.84 & 2.25 \\ \hline
TEMPGONE & 3.95 & 9.62 & 3.62 & 7.28 & 3.88 & 3.06 \\ \hline
TEMPNITE & 3.41 & 9.05 & 3.11 & 7.04 & 3.24 & 2.96 \\ \hline
USEWWAC & 0.28 & 1.08 & 0.3 & 1.03 & 0.31 & \textbf{0.26} \\ \hline
TEMPHOMEAC & 2.43 & 18.46 & 2.09 & 12.97 & 3.67 & 1.99 \\ \hline
TEMPGONEAC & 3.22 & 20.63 & 3.08 & 66.86 & 11.19 & 2.74 \\ \hline
TEMPNITEAC & 2.84 & 17.56 & 2.35 & 12.73 & 3.66 & 2.31 \\ \hline
HHAGE & 0.41 & 0.74 & 0.62 & 0.72 & 0.45 & 0.39 \\ \hline
EMPLOYHH & 0.38 & 0.56 & 0.44 & 0.53 & 0.33 & 0.33 \\ \hline
EDUCATION & 0.78 & 1.12 & 1.02 & 0.9 & 0.85 & 0.77 \\ \hline
NHSLDMEM & \textbf{0.0009} & 0.72 & 0.02 & 1.04 & 1.02 & 0.35 \\ \hline
NUMADULT & \textbf{0} & 0.46 & 0.01 & 0.56 & 0.33 & 0.32 \\ \hline
NUMCHILD & \textbf{0.002} & 0.53 & 0.02 & 0.58 & 0.52 & 0.37 \\ \hline
ATHOME & 1.57 & 1.86 & 1.62 & 1.53 & 1.4 & 1.47 \\ \hline
MONEYPY & 1.3 & 2 & 1.64 & 2.02 & 1.31 & 1.38
\end{tabular}
\label{tab:test_mae}
\end{table}

Mean Absolute Error (MAE) is one important metric for summarizing and assessing the quality of a machine learning model. MAE refers to the mean of the absolute values of each prediction error on all instances of the testing dataset. The prediction error is the difference between the actual value and the predicted value for that instance. MAE can be calculated using the following formula: 

\begin{equation}
\int_{i=1}^n {\frac{abs(y_i - \lambda(x_i))}{n}}
\label{eq:mae_eq}
\end{equation}

where $n$ is the number of test samples, $y_i$ is actual class label for $x_i$ input and  $\lambda(x_i)$ is the predicted class label. Table \ref{tab:test_mae} illustrates the MAE of the trained classification models in the test suite. The target variable TEMPGONEAC has the highest MAE ($66.86$) for Support Vector Machine classifier (SVM), and the target variable NUMADULT has the lowest MAE ($0.01$) for Decision Tree Classifier (CART). Except for the highest occurrence of the MAE for TEMPGONEAC, the MAE scores of different models indicate similar characteristics over different target variables. The Random Forest Classifier (RFC) maintained the lowest MAE score for most targets with an average of $1.96$. 

\begin{table}[]
\caption{$R^2$ score in test set}
\begin{tabular}{lllllll}
 & LDA & KNN & CART & SVM & ADB & RFC \\ \hline
EQUIPMUSE & 0.42 & -0.31 & 0.32 & -0.4 & 0.28 & 0.42 \\ \hline
TEMPHOME & 0.92 & -0.27 & 0.94 & -0.04 & 0.93 & \textbf{0.95} \\ \hline
TEMPGONE & 0.86 & -0.35 & 0.88 & -0.07 & 0.88 & \textbf{0.91} \\ \hline
TEMPNITE & 0.9 & -0.3 & 0.9 & -0.04 & 0.91 & \textbf{0.92} \\ \hline
USEWWAC & \textbf{0.86} & -0.29 & 0.81 & -0.25 & 0.8 & 0.83 \\ \hline
TEMPHOMEAC & 0.98 & -0.77 & 0.97 & -0.07 & 0.96 & \textbf{0.98} \\ \hline
TEMPGONEAC & 0.97 & -0.89 & 0.96 & -6.48 & 0.33 & \textbf{0.97} \\ \hline
TEMPNITEAC & 0.97 & -0.63 & 0.97 & -0.09 & 0.95 & \textbf{0.98} \\ \hline
HHAGE & 0.43 & -0.36 & -0.09 & -0.2 & 0.3 & 0.41 \\ \hline
EMPLOYHH & -0.2 & -0.5 & -0.39 & -0.21 & 0.002 & 0.03 \\ \hline
EDUCATION & 0.16 & -0.53 & -0.38 & -0.01 & -0.003 & 0.18 \\ \hline
NHSLDMEM & 0.9995 & 0.27 & 0.97 & -0.16 & -0.09 & 0.64 \\ \hline
NUMADULT & 1 & 0.07 & 0.97 & -9.4 & 0.27 & 0.32 \\ \hline
NUMCHILD & 0.998 & -0.089 & 0.96 & -0.32 & 0.01 & 0.28 \\ \hline
ATHOME & -0.61 & -0.96 & -0.53 & -0.61 & \textbf{-0.42} & -0.53 \\ \hline
MONEYPY & 0.32 & -0.44 & -0.06 & -0.57 & 0.25 & 0.17
\end{tabular}
\label{tab:test_r2}
\end{table}

The R-squared ($R^2$) score is another important metric that is used to evaluate the performance of machine learning models. $R^2$ type goodness-of-fit summary statistics have been constructed for particular models using a variety of methods \cite{cameron1997r}. It works by measuring the amount of variance in the predictions explained by the dataset and represents the difference between the samples in the dataset and the predictions made by the model. $R^2$ score can be calculated using the following formula: 

\begin{equation}
R^2(y, \hat{y}) = 1 - \frac{\sum_{i=1}^n (y_i-\hat{y_i})^2}{\sum_{i=1}^n (y_i-\bar{y_i})^2}
\label{eq:r2}
\end{equation}

where $y$ is the actual class, $\hat{y}$ is predicted class and $\bar{y}$ is the mean. This basically means:

\begin{equation*}
R^2 = \frac{Explained Variation}{Total Variation}
\end{equation*}

If the value of the $R^2$ is 1, it means that the model is perfect, and if its value is 0, it means that the model will perform badly on an unseen dataset. The machine learning models chosen in this task have mixed results in $R^2$ score. Some of the models performed very well for some targets like LDA having $R^2$ score of $1$ for NUMADULT, $0.998$ for NUMCHILD, and $0.9995$ for NHSLDMEM. Also, some of the models triggered poor results such as SVM having a negative $R^2$ score for all targets. 


\section{Discussion}

\subsection{Findings}
\label{sec:findigns}

\begin{table}[]
\caption{Best scores in test set}
\begin{adjustbox}{width=\textwidth/2,center}
\label{tab:best_test_scores}
\begin{tabular}{llll}
\textbf{Target} & \textbf{Test Accuracy} & \textbf{MAE} & \textbf{R2} \\ \hline
EQUIPMUSE & 0.67 (RFC) & 0.57 (RFC) & 0.42 (RFC) \\ \hline
TEMPHOME & 0.43 (CART) & 1.84 (ADB) & 0.95 (ADB) \\ \hline
TEMPGONE & 0.38 (CART) & \textbf{3.06} (RFC) & 0.91 (RFC) \\ \hline
TEMPNITE & 0.36 (CART) & 2.96 (RFC) & 0.92 (RFC) \\ \hline
USEWWAC & \textbf{0.89} (RFC) & 0.26 (RFC) & 0.86 (LDA) \\ \hline
TEMPHOMEAC & 0.53 (CART) & 1.99 (RFC) & 0.98 (RFC) \\ \hline
TEMPGONEAC & 0.42 (CART) & 2.74 (RFC) & 0.97 (RFC) \\ \hline
TEMPNITEAC & 0.46 (CART) & 2.31 (RFC) & 0.98 (RFC) \\ \hline
HHAGE & 0.64 (RFC) & 0.39 (RFC) & 0.43 (RFC) \\ \hline
EMPLOYHH & 0.73 (ADB) & 0.33 (ADB, RFC) & \textbf{0.03} (ADB, RFC) \\ \hline
EDUCATION & 0.39 (RFC) & 0.77 (RFC) & 0.18 (RFC) \\ \hline
NHSLDMEM & \textbf{0.9991} (LDA) & 0.0009 (LDA) & 0.9995 (LDA) \\ \hline
NUMADULT & \textbf{1} (LDA) & 0 (LDA) & \textbf{1} (LDA) \\ \hline
NUMCHILD & \textbf{0.998} (LDA) & 0.002 (LDA) & 0.998 (LDA) \\ \hline
ATHOME & 0.57 (SVM, ADB, RFC) & 1.4 (ADB) & \textbf{-0.42} (ADB) \\ \hline
MONEYPY & 0.34 (ADB) & 1.3 (LDA) & 0.32 (LDA) \\ \hline
\textbf{Average} & \textbf{0.612944} & \textbf{1.245181} & \textbf{0.657969}
\end{tabular}
\end{adjustbox}
\end{table}

Several findings can be deduced by analyzing the results in sections \ref{sec:results}. First, the accuracy obtained by the machine learning models is almost $99\%$ for some of the target variables such as the number of household members (NHSLDMEM), the number of adults (NUMADULT), and the number of children (NUMCHILD) living in the household, whereas an average of $37\%$ for some other target variables such as the winter temperature when no ones at home (TEMPGONE), the winter temperature at night (TEMPNITE), the highest level of education obtained by the respondent (EDUCATION), and the annual gross household income (MONEYPY). The MAE and $R^2$ scores follow similarly in these target variables with MAE being close to $0$ and $R^2$ score close to $0.99$ for NHSLDMEM, NUMADULT, and NUMCHILD. The MAE ranges between $1$ to $3$ for TEMPGONE, TEMPNITE, EDUCATION, and MONEYPY. The $R^2$ score looks promising for TEMPGONE and TEMPNITE (around $0.92$) but not as good in the case of EDUCATION ($0.18$) and MONEYPY ($0.32$). The target variable representing the behavior of the most used individual air conditioning unit (USEWWAC) resulted constructive with an accuracy of $89\%$, a MAE of $0.26$, and a $R^2$ score of $0.86$. Overall, the $R^2$ score remains above $0.80$ for 10 target variables, denoting the machine learning models are performing well for these target variables. Only the target variable representing the number of weekdays someone is present at home (ATHOME) achieved a $R^2$ score of $-0.42$. The results indicate clearly that the models can predict NHSLDMEM, NUMADULT, NUMCHILD, and USEWWAC with very high confidence. The overall performance of the machine learning models on the target variables EDUCATION and MONEYPY imply that the behavior of occupants with different educational backgrounds and income levels are not that different.
\\

Table \ref{tab:best_test_scores} also shows that the decision tree classifier (CART) achieves best accuracy for most of the targets whereas the random forest classifier achieves the best scores in terms of MAE and $R^2$ for most of the target variables. Overall, the average accuracy, mean absolute error, and $R^2$ score remains $61\%$, $1.24$, and $0.66$, respectively through all $16$ target variables. It proves that the machine learning models can be used with a certain confidence to automate some steps in the building occupant persona generation task. This also proves that there is still room for improvement in the area. The machine learning models are kept in their default settings without any hyper-parameter tuning, which can be one of the reasons for the lower accuracy results for some of the target variables. The evaluation results of the models recapitulate the following findings: 

\begin{itemize}
    \item It is possible to use machine learning tools to semi-automate building occupant persona development. 
    \item With sufficient training data, the machine learning models can produce $90\%$ and above accuracy in some occupant characteristics prediction.
    \item Some occupant characteristics are independent of occupant behavior and energy consumption properties. 
\end{itemize}
 
Agee \textit{et al}. \cite{agee2021human} presented a human-centered smart housing development process involving multiple steps of data analysis, visualization, and classification that involve human supervision. After the energy analysis, behavioral analysis, semi-structured interviews, and affinity diagramming, the authors synthesized the results to produce a senior persona and a non-senior persona. The developed persona incorporates information such as age, physical needs, temperature controls, usage of electronic appliances, e.g., television, dishwasher, behavior, etc. The following is a snippet of the provided persona example:
\\

\textit{``Inez is a 77-year-old retiree. She lives by herself, but keeps herself busy with her church group, visiting grandkids, reading, and TV. She lives alone, so feeling safe and secure is important. She sets her thermostat between $72-75^{\circ}F$. She uses $299 kBtu/m^2/yr$ of energy. She has an energy start dishwasher but prefers washing by hand. She can't afford to be wasteful, and the monthly energy bill is hard to understand.''}
\\

This work incorporates $16$ target variables that are useful in persona development, for example, HHAGE predicts occupant age, EDUCATION provides the level of education of the occupant, EQUIPMUSE provides the behavior of the main heating equipment in the household, TEMPHOME provides winter temperature when someone is home during the day, and TEMPHOMEAC illuminates summer temperature when someone is home during the day. Section \ref{sec:target_variables} discusses the complete details of each target variable. With the prediction results of all target variables, the following characteristics of a building occupant persona can be produced automatically: (i) number of people living in the household, (ii) number of adults living in the household, (iii) number of children living in the household, (iv) preferred temperature during winter, (v) preferred temperature during summer, (vi) the primary usage of air conditioner during summer, (vii) behavior of the heater during winter, (viii) the age of the respondent, and (ix) the employment status of the respondent. In the smart housing persona development process proposed by \cite{agee2021human}, labeling and classification of these characteristics are done manually. The results of this work prove that these steps can be automatically completed by machine learning models and thus reduce manual workload significantly in the process of developing building occupant personas.  

\subsection{Answer to the research questions}
\begin{itemize}
    \item RQ1: Is it possible to automate the process or some steps of building occupant persona development?
    \\
    It is definitely possible to automate multiple steps of the smart persona development process, as shown in section \ref{sec:findigns}. The process of smart housing persona development requires manual human tasks for filtering, clustering, and labeling collected datasets. Going through each record in a large dataset can be extensively wearisome. These steps can be automated with machine learning tools as demonstrated in this work. The selection of target variables in the dataset determines the level of detail the machine learning models may produce. Section \ref{sec:methodology} illustrates the process of smart housing persona characteristics prediction from the residential energy consumption dataset. Although this process is not fully autonomous, it eliminates manual human work on steps such as filtering, clustering, and labeling, which accelerates the entire process of smart housing persona generation. Table \ref{tab:best_test_scores} shows the best accuracy, MAE and $R^2$ score for all the target variables in the RECS 2015 dataset. This answers the question of using similar datasets for the same task. Although it cannot be guaranteed that the models will perform similarly in other datasets, it can be assumed that the proposed method and process will produce results comparable to this work on similar datasets.  
    \\
    \item RQ2: How accurate results can be achieved in the automated process of building occupant persona development? 
    \\
    The answer to this question lies in the machine learning model selection and training process. A precisely chosen and configured model can predict results with near $100\%$ accuracy, such as LDA on the number of household members (NHSLDMEM), the number of adults (NUMADULT), and the number of children (NUMCHILD) living in the household. Whereas a randomly chosen model might provide accuracy as low as $11\%$, for example, the accuracy of the winter temperature when no one is at home (TEMPGONE), the winter temperature at night (TEMPNITE), the summer temperature when someone is present at home (TEMPHOMEAC), and the summer temperature when no one is at home (TEMPGONEAC). This work demonstrates both scenarios. The outcome of a machine learning model is highly dependent on the dataset it is trained on. The models in this work are trained on only one dataset with $5686$ records and they show promising results. Introducing more data will further evolve the model's prediction capabilities and improve its performance. Hyper-parameter tuning can also add to that performance. Table \ref{tab:best_test_scores} shows that the models achieved $89\%$ and above in terms of accuracy for the behavior of the most used individual air conditioning unit (USEWWAC), the number of household members (NHSLDMEM), the number of adults (NUMADULT), and the number of children (NUMCHILD) living in the household. Target variables referring to the behavior of the main heating equipment (EQUIPMUSE), the age of the respondent (HHAGE), and the employment status of the respondent (EMPLOYHH) achieved $64\%$ and above in terms of accuracy. The $R^2$ scores noted in table \ref{tab:best_test_scores} shows that the models achieve a score of $0.90$ and above for $10$ of the target variables. Overall, the machine learning models used in this task achieve an average accuracy of $61\%$, an average MAE of $1.24$, and an average $R^2$ score of $0.66$. The best accuracy, MAE, and $R^2$ score for all the target variables are shown in table \ref{tab:best_test_scores} including the names of the best performing models. 
\end{itemize}

\subsection{Limitations and future research}
This research automates multiple steps of the smart housing persona generation process to reduce manual labor and save time. Nonetheless, it does not fully automate the task. There are some limitations that can be resolved or mitigated in the future. The accuracy of the machine learning models is not up to the mark for all the target variables. Even though the models perform well for many of the targets, for some of them, the models perform below an accuracy of $60\%$. The models are trained on one dataset only, and the generalizability of the proposed approach has not been tested on a different dataset. We plan to apply the entire approach to the RECS 2020, which is set to release on 2023 \cite{eia_2023}. 
\\

Future work will include hyper parameter tuning of the machine learning models for improved accuracy, and testing other machine learning models. Developing deep learning models for occupant persona generation can also be useful because deep learning models are proven to achieve better accuracy for dedicated tasks compared to generic machine learning models. Future research can utilize all this work to fully automate the persona development process, and thus, promote functions and accuracy of building performance simulation, design behavior interventions, and develop smart building management solutions in a more holistic way.


\section{Conclusion}
This research investigates the feasibility of using machine learning to automate some steps of building persona development. Five types of machine learning models are developed using the 2015 Residential Energy Consumption Survey data provided by U.S. Energy Information Administration. The results indicate that it is possible to use machine learning tools to partially automate the process of building occupant persona development. This research contributes to the body of knowledge by proposing a machine learning based approach for facilitating the development of building occupant personas. It proves that given the necessary data, it is possible to automatically predict the occupant characteristics with a certain confidence, and thus, help us to better understand occupants and improve living conditions to meet occupant demands.


\bibliographystyle{IEEEtran}
\bibliography{bare_conf}

\end{document}